\def\year{2016}\relax
\documentclass[letterpaper]{article}

\usepackage[utf8]{inputenc}
\usepackage{url}
\usepackage{graphicx}
\usepackage{subfigure}
\usepackage{tabularx}
\usepackage{multirow, array}

\newcommand{\citeonline}[1]{\citeauthor{#1} (\citeyear{#1})}
\newcommand{\cellvalign}[1]{\vspace{0.01pt} #1}

\usepackage{arxiv16}
\usepackage{times}
\usepackage{helvet}
\usepackage{courier}
\begin{document}
%
\title{Fusing Audio, Textual and Visual Features for Sentiment Analysis of News Videos$^*$}
\author{Moisés H. R. Pereira$^1$, Flávio L. C. Pádua$^2$, Adriano C. M. Pereira$^3$,\\
	{\bf \Large Fabrício Benevenuto$^4$, Daniel H. Dalip$^5$}\\
	$^{1,5}$Engineering and Technology Institute, University Center of Belo Horizonte (UNI-BH), Belo Horizonte, MG, Brazil\\
	$^2$Department of Computing, CEFET-MG, Belo Horizonte, MG, Brazil\\
	$^{3,4}$Department of Computer Science, Federal University of Minas Gerais (UFMG), Belo Horizonte, MG, Brazil\\
	moises.ramos@prof.unibh.br, cardeal@decom.cefetmg.br, \{adrianoc, fabricio\}@dcc.ufmg.br, dhasan@prof.unibh.br
}
\maketitle
\begin{abstract}
This paper presents a novel approach to perform sentiment analysis of news videos, based on the fusion of audio, textual and visual clues extracted from their contents. The proposed approach aims at contributing to the semiodiscoursive study regarding the construction of the ethos (identity) of this media universe, which has become a central part of the modern-day lives of millions of people. To achieve this goal, we apply state-of-the-art computational methods for (1) automatic emotion recognition from facial expressions, (2) extraction of modulations in the participants' speeches and (3) sentiment analysis from the closed caption associated to the videos of interest. More specifically, we compute features, such as, visual intensities of recognized emotions, field sizes of participants, voicing probability, sound loudness, speech fundamental frequencies and the sentiment scores (polarities) from text sentences in the closed caption. Experimental results with a dataset containing 520 annotated news videos from three Brazilian and one American popular TV newscasts show that our approach achieves an accuracy of up to 84\% in the sentiments (tension levels) classification task, thus demonstrating its high potential to be used by media analysts in several applications, especially, in the journalistic domain. 
\end{abstract}

\section{Introduction}

The newscast is the backbone of all television networks in the world. In order to transmit credibility, during a TV newscast, it can be often seen a spectacularization of the information by the journalists as a way to prepare the mental model of the viewers~\cite{Goffman:1981}. 

As the enunciative communication process is singular (i.e. made usually with just one interlocutor), the information subject choose the style resources in which he/she will use for an specific communication situation such as vocal modulations, corporal expressivity, allowing the newscast to stand on what is being reported and to establish an identification with the viewer~\cite{EisensteinBarzilayDavis:2008}. In addition, TV newscasts show the story tags followed by some news with high or moderated emotional content, varying the emotional tension as more news are presented during the program. Generally, the last news have  content with low emotional tension, except in moments of great social upheaval~\cite{PereiraPaduaDavid-Silva:2015}.

Newscast analysis is essential to media annalists in several domains, especially in the journalism field~\cite{Stegmeier:2012}. Since newscasts is a specific type of discourse and a sociocultural practice, discourse analysis techniques~\cite{Charaudeau:2002} have been applied in order to analyse the newscast structure in many levels of descriptions, concerning some properties, such as their general thematics, enunciation schemes and discourse style as newscast production dimensions~\cite{Cheng:2012}.

Normally, speeches are analysed without the support of computational tools such as automatized annotation softwares and video analytic programs. Only recently, with the development of some areas such as the sentiment analysis, computational linguistics, multimedia systems and computer vision, new methods have been proposed to support the discourse analysis, especially in multimedia content such as TV newscasts~\cite{Pantti:2010}. However, to the best of our knowledge, there is no previous effort that has attempted to use multimodal features (e.g. audio, textual and visual features) in order to measure the tension of the news. Also, approaches to infer the tension can have a good applicability to infer importance of the news and then helping to organize and summarize news.

In a first step towards this goal, we present here a computational approach in order to support the study of tension level in news from multimodals features available in their videos. The proposed approach allows researchers to perform a semiodiscursive analysis of verbal and non-verbal languages which are manifested through facial expressions and by gestural movements of the journalists in front of the camera which is a visual way to express their ideas~\cite{EisensteinBarzilayDavis:2008}. In our experiments, we shown the effectiveness of proposed approach as well as the importance of sentiment analysis to infer high tension news.

\section{Related Work}

\subsection{Tension and Sentiment Analysis in News}

Many people read online news from websites of the great communication portals. These news websites need to create effective strategies to draw people attention to these contents. In this context, \citeonline{Reis.etal:2015} investigate strategies used by online news organizations in designing their headlines. It was analysed the content of 69,907 headlines produced by four major media companies during a minimum of eight consecutive months in 2014. They find out that news with a negative sentiment tend to generate many views as well as negative comments. In the result analysis, authors point out that the greater the negative tension of the news, the greater the need that the user feels to give their opinion. This, in controversial subjects, has a great possibility of being a comment that contradicts the opinion of another user, promoting discussions on posts.

\citeonline{Pantti:2010} studies the emotional value of the journalists expressions in newscasts at the expense of public emotions. This work provides evidences of how the journalists evaluate the role and position of emotions in the media coverage and the empathy induced by the news. Furthermore, this work studies how the journalists emotional discourse is linked to their idea of good journalism and their professional image. To accomplish this, they used a set of interviews from television professionals who work with utility programs and advertising news from Finland and Netherlands.

\subsection{Multimodal Sentiment Analysis}

Regarding multimedia files, it is not enough to process only one information modality in order to assertively perform sentiment analysis in this content. In this context, \citeonline{Poria.etal:2016} presented an innovative approach for multimodal sentiment analysis which consists of collecting sentiment of videos on the Web through a model that fuses audio, visual and textual modalities as information resources. A feature vector were created in an fusion approach in an feature and decision level, obtaining a precision of around 80\%, representing a increase of more than 20\% of precision when comparing to all the state-of-the-art systems. 

\citeonline{MaynardDupplawHare:2013} describe an approach for sentiment analysis based on social media content, combining opinion mining in text and in multimedia resources (e.g. images, videos),  focusing on entity and event recognition to help archivists in selecting material for inclusion in social media in order to solve ambiguity and to provide more contextual information. They use Natural Language Processing (NLP) tools and a rule based approach for the text, concerning issues inherent to social media, such as grammatically incorrect text, use of profanity and sarcasm.

\subsection{Audio and Video Emotion Recognition}

\citeonline{EkmanFriesen:1978} showed evidences that facial expressions of emotions can be inferred by rapid face signs. These signals are characterized by changes in face appearance that last seconds or fractions of a second. Thus, the authors formulated a model of basic emotions, called the Facial Action Coding System (FACS), based on six facial expressions (happiness, surprise, disgust, anger, fear and sadness) which are found in many cultures and presented in the same way, from children to seniors. The singularity points were mapped to each type of facial expression through tests on a vast image database.

Regarding the recognition of prosodic features in the sound modulations of audio signals, \citeonline{Eyben.etal:2013} showed the development of openSMILE, a framework for extracting features of emotional speech, music and sounds in general from videos and audio signals. The activity detection, voice monitoring and face detection are also resources offered by this framework.

We can observe that there is a demand in analyzing the emotional content of videos and news in many types of media. In this context, this paper applies robust techniques that allow to automatically  determine the emotional tension levels using sentiment analysis techniques in news videos for content-based analysis of news programs.

\section{The Proposed Approach}

This section presents the multimodal features used and our approach to combine textual, visual and audio information in order to perform the computation of tension levels in the narrative content from events shown in the videos.

\subsection{Multimodal Features}

In this work, multimodal features are organized into two groups: (1) the audiovisual clues and (2) the sentiment scores of textual sentences obtained from the closed caption (textual information). Audio visual clues are represented by the visual intensity of the recognized emotion, the participants field size and the prosodic features of the audio signal that corresponds at aspects of speech that go beyond the phonemes and deal with the sound quality: voicing probability, loudness and fundamental frequency.

\subsubsection{Visual Intensity} This feature was measured based on the output margin, in other words, the distance to the separation hyperplane of the classifiers used in \citeonline{Bartlett.etal:2006} under a one-to-all approach, where it was used a Linear SVM classifier for each of the modeled emotions (happiness, surprise, aversion, contempt, anger, fear and sadness). The visual intensity of the emotion is computed for each frame of the news video under analysis. In frames where no face was detected, the system computes the emotion as \textit{Nonexistent}.

\subsubsection{Participants Field Size} We calculate the ratio between the recognition area of the largest face detected in the frame and the current frame resolution to extract the proportion of faces values of the participants in the camera placement. In the TV newscast universe, when there are several individuals in different field sizes of camera placement, the focus will be on the individual with the closest field size, that is, whose face occupies the largest area of the frame. Furthermore, the larger the area that the face occupies in the frame, the greater the emotional intensity that it was pretended to be applied as a communication strategy of the program in the varied use of field size during the exhibition \cite{Gutmann:2012}.

\subsubsection{Voicing Probability} Shows the probability to have a tone differentiation during the speech in the next moment.

\subsubsection{Sound Loudness} Sound loudness reflects the perception of loudness of the sound wave by the human ear measured in decibels (dB).

\subsubsection{Fundamental Frequency} Corresponds to the first harmonic of a sound wave, this is the most influential frequency for the perception of a particular sound and one of the main elements for characterizing the voice~\cite{Eyben.etal:2013}.

\subsubsection{Sentiment Scores} Extracted from the closed caption. Each sentence is formed from a subtitle text and it is analyzed by 18 state-of-the-art methods \cite{Araujo.etal:2014} for textual sentiment analysis, generating a vector of 18 scores (-1, 0, or +1), one for each method. We sum its values in order to obtain the sentiment score for that sentence.

\subsection{Tension Levels Computation}

\begin{figure}[!b]
	\centering
	\includegraphics[scale=0.50]{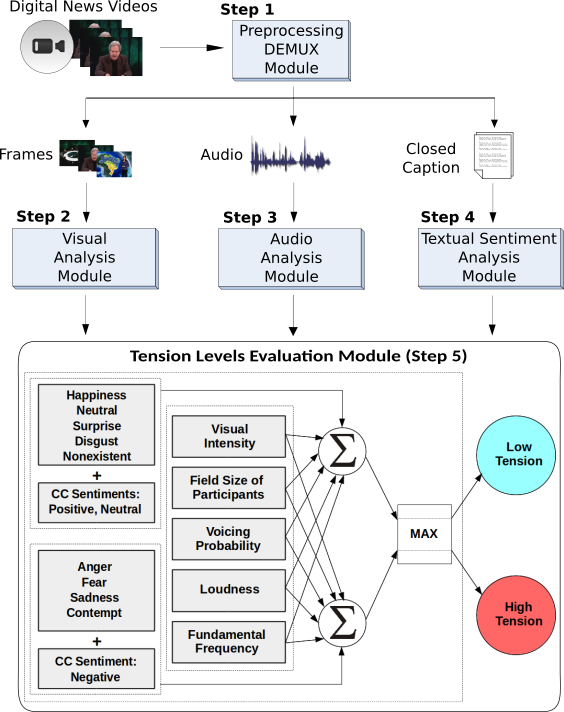} 
	\caption{Overview of the proposed approach in sentiment analysis for tension levels computation.} 
	\label{TensionLevelsComputation}
\end{figure}

In this paper, we extract data from the (1) audio prosodic features, (2) the visual features in the videos, (3) and the sentiment scores extracted on the closed caption. After that, we determine the corresponding tension levels. Figure \ref{TensionLevelsComputation} presents the overview of the proposed approach.

In Step 1, the developed system extracts the multimodal resources from TV news videos, where, each video contains different news. By doing this, we can obtain a list of images from the video (frames), the audio signal in WAV format and the text obtained from the closed caption. These modalities of visual, audio and textual information are organized in three processing lines, one for each modality.

In Step 2, using the obtained frames from the previous step, we apply the methods of facial expressions and emotion recognition proposed in \citeonline{Bartlett.etal:2006}, obtaining the values for the visual features (visual intensity and field size) on the emotional expressiveness of the face.

In Step 3, the audio signal is processed by extracting the corresponding acoustic data, also considering the speech instants of individuals, that is, when speech occurs. To accomplish this, we use the openSMILE framework \cite{Eyben.etal:2013} to extract the audio component of the spectrum and obtain the prosodic features of loudness, voicing probability and the fundamental frequency of the speech modulations for each hundredth of a second these audio signals.

In Step 4, the closed caption is processed in order to extract only the textual content regarding the transcribed speech (subtitle text). After that, each subtitle text is transformed into a single sentence. Each sentence is recorded in a text file and then the text is submitted to sentiment analysis process. To achieve this goal, we used the iFeel to extract the sentiment polarities of each sentence, classifying them as positive, neutral or negative, for the 18 state-of-the-art methods \cite{Araujo.etal:2014}. Note that, the emoticons were not used since in the closed caption there are not the specific character set used to represent emotions, proposed the emoticons method~\cite{Park.etal:2013}.

The Step 5, shown in Figure \ref{TensionLevelsComputation}, obtains the tension level of a specific video based on the sum with the highest value among the two tension levels mapped: Low Tension and High Tension. The sums of the emotions recognized by the facial expressions and sentiment scores of the closed captions for each tension level are weighted by audiovisual clues calculated over the video.

A particular news video is classified by the tension level whose emotions were totaled with the highest values of multimodal features calculated throughout the video during the automatic recognition.

\section{Preliminary Results}

In this section we describe the dataset, the annotation process, the performed experiments on the implemented system and discusses the preliminary results in order to evaluate the performance of the proposed approach in some aspects.

\subsection{Dataset and Annotation Process}

The dataset constructed to evaluate the proposed approach in this work has 520  news videos obtained from 27 exhibitions of four TV newscasts, three Brazilian news programs (namely \textit{Jornal da Record}, \textit{Band News} and \textit{Jornal Nacional}) and an American television news program (CNN)\footnote{Among the 520 videos of the dataset, 264 have closed caption. Then, in videos which does not have closed caption, they received zero in this feature.}. More specifically, we used  226 news videos from \textit{Band News} shown from 10th to 14th of June, 2013, from 23rd to 28th of April and from 7th to 12nd of December, 2015; 237 videos of the \textit{Jornal da Record} (JR) shown on May 24th, 2013, January 10th, 2015, February 5th, 2015 and 2nd 10th 16th and 18th of March, 2015; 47 videos from \textit{Jornal Nacional} (JN) shown on January 20th and December 17th,  2015; and 10 videos of CNN News shown on April 6th, 2015.

In order to evaluate the proposed model, four professionals from several fields of knowledge (Journalism and Languages, Mathematical Modeling, Law and Pedagogy) made manual annotations of the tension levels of the news videos. The tension levels considered for annotation and explained below were Low Tension and High Tension. Among the 520 videos analyzed, all contributors have annotated the same tension level to 381 videos, thus reaching 73.27\% of agreement and in 18.46\% of the annotations (96) reached an agreement among 75\% of the contributors. 

\subsection{Field Sizes x Sentiment Scores x Our Approach}

We analyze  the performance of our newly proposed features for inferring the tension level of newscast and also our proposed approach. Furthermore, we performed our comparison in two manners: (1) considering all the videos and (2) considering only those videos that all the annotators agreed each other (100 \% of annotation concordance).  

Table~\ref{tab:fieldSizesXSentimentScores} shows the accuracy considering this two approaches when using just specific features and using our approach. As expected, the performance was better when there was 100\% agreement among the annotators. Note that, to compare the approaches, we performed paired t-test between the processing approaches and the results are different with an 95\% of acuracy.

\begin{small}
\begin{table}[t]
	\centering
	\begin{tabularx}{0.45\textwidth}{p{2.0cm}|p{2.3cm}|p{2.4cm}}
		\hline
		\textbf{Some} & \multicolumn{2}{c}{\textbf{Processing Approach}}\\
		\cline{2-3}
		\textbf{Multimodal Features} & \textbf{All \newline Annotations} & \textbf{100\% of \newline Concordance}\\
		\hline
		\textbf{Participants Field Sizes}  &
		\cellvalign{0.79} &
		\cellvalign{0.85}\\
		\hline
		\textbf{Sentiment Scores} & 
		\cellvalign{0.72} &
		\cellvalign{0.83}\\     
		\hline
		\textbf{Proposed Method} & 
		\cellvalign{0.78} &
		\cellvalign{0.84}\\   
		\hline
	\end{tabularx}
	\caption{Acuracy for each processing approach for the newly proposed multimodal features and our proposed approach.}
	\label{tab:fieldSizesXSentimentScores}
\end{table}
\end{small}

In the experiments implemented, we could observe that sentiment scores were the best for inferring High Tension videos. Thus, we can conclude that Sentiment Scores are good to infer High Tension videos, however, Sentiment Scores alone are not the best to infer Lower Tension videos. Then, if we want a better performance for the Low Tension videos, we can use our proposed approach.

\section{Concluding Remarks}

Facial expressions are forms of non-verbal communication that externalize our demonstrations of stimulus in which we are subjected and, therefore, are elements that are part of media news. In order to legitimize the reported fact as communicative strategy, newscasters express their emotions, providing evidences about tension of the speech generated by the news and the pattern in the sequencing of the news contained in the program. In this sense, this work presented one approach to infer the tension of news videos taking into account multiple sources of evidences.

By experiments we could show that sentiment analysis of the text, which were for the first time used in this domain, can be important for inferring high tension news. We also show that our approach can have a good performance to infer High tension news. However, for the identification of Low Tension news, our approach are only worse than using just sentiment analysis.

\section{Acknowledgments}
The authors would like to thank the support of CNPq-Brazil under Procs. 468042/2014-8 and 313163/2014-6, FAPEMIG-Brazil under Proc. APQ-01180-10, CEFET-MG under Proc. PROPESQ-023-076/09, and of CAPES-Brazil.

\bibliography{references}
\bibliographystyle{aaai}

\end{document}